\documentclass{article}

\usepackage{microtype}
\usepackage{graphicx}
\usepackage{subfigure}
\usepackage{booktabs} 

\usepackage{hyperref}

\usepackage[accepted]{icml2021}

\usepackage[utf8]{inputenc} 
\usepackage[T1]{fontenc}    
\usepackage{hyperref}       
\usepackage{url}            
\usepackage{booktabs}       
\usepackage{amsfonts}       
\usepackage{nicefrac}       
\usepackage{microtype}      
\usepackage{multirow}
\usepackage{stmaryrd }
\usepackage{amssymb }
\usepackage{amsmath }
\usepackage{amsfonts}
\usepackage{dsfont}
\usepackage{amsthm}

\usepackage{enumerate}

\newcommand{\R}{\mathbb{R}}

\newcommand{\bra}{\langle}
\newcommand{\ket}{\rangle}

\newcommand{\by}{\bold y}

\newcommand{\bPhi}{\boldsymbol \Phi}
\newcommand{\bvarphi}{\boldsymbol \varphi}

\newcommand{\bSigma}{\boldsymbol \Sigma}
\newcommand{\bC}{{\bold C}}
\newcommand{\bCi}{{\bold C^{-1}}}
\newcommand{\bmu}{{\boldsymbol \mu}}

\newcommand{\bGamma}{\boldsymbol \Gamma}
\newcommand{\bgamma}{\boldsymbol \gamma}

\newcommand{\act}{\mathcal{A}} 
\newcommand{\supp}{\mathcal{S}} 

\newcommand{\beps}{{\boldsymbol \epsilon}}

\newcommand{\sub}[1]{_{\text{\tiny #1}}}

\newcommand{\N}{\mathcal{N}}

\newcommand{\eqdef}{\overset{\text{\tiny{def}}}{=}}

\newcommand{\RMP}{RMP$_\sigma$}
\newcommand{\RMPZ}{RMP$_0$}

\newtheorem{thm}{Theorem}
\newtheorem{defi}[thm]{Definition}
 \newtheorem{lem}[thm]{Lemma}
\newtheorem{cor}[thm]{Corollary}

\numberwithin{thm}{section}
\numberwithin{equation}{section}

\DeclareMathOperator\erf{erf}

\usepackage{multicol}
\usepackage{graphicx}

\icmltitlerunning{Sparse Bayesian Learning via Stepwise Regression}

\begin{document}

\twocolumn[
\icmltitle{Sparse Bayesian Learning via Stepwise Regression}

\begin{icmlauthorlist}
\hspace*{\fill}
\icmlauthor{Sebastian Ament}{cornell}
\hfill
\icmlauthor{Carla Gomes}{cornell}
\hspace*{\fill}
\end{icmlauthorlist}
\icmlaffiliation{cornell}{Department of Computer Science, Cornell University, Ithaca, NY, USA}
\icmlcorrespondingauthor{Sebastian Ament}{ament@cs.cornell.edu}
\icmlkeywords{Sparse Bayesian Learning, Stepwise Regression, Compressed Sensing, Feature Selection}
\vskip 0.3in
]

\printAffiliationsAndNotice{}  

\begin{abstract}
Sparse Bayesian Learning (SBL) is a powerful framework for attaining sparsity in probabilistic models.
Herein, we propose a coordinate ascent algorithm for SBL
termed Relevance Matching Pursuit (RMP) and show that,
as its noise variance parameter goes to zero,
RMP exhibits a surprising connection to Stepwise Regression.
Further, we derive novel guarantees for Stepwise Regression algorithms,
which also shed light on RMP.
Our guarantees for Forward Regression improve on deterministic and probabilistic results for Orthogonal Matching Pursuit with noise. 
Our analysis of Backward Regression on determined systems 
culminates in a bound on the residual of the optimal solution to the subset selection problem
that, if satisfied, guarantees the optimality of the result.
To our knowledge, this bound is the first that can be computed in polynomial time and depends chiefly on the smallest singular value of the matrix. 
We report numerical experiments using a variety of feature selection algorithms.
Notably, RMP and its limiting variant are both efficient and maintain strong performance with correlated features.
\end{abstract}


\section{Introduction}
Finding a sparse solution to underdetermined linear systems is a fundamental problem in a diverse array of domains and applications, like network systems
\citep{networksystems}, materials science \citep{szameit2012sparsity}, medical imaging \citep{lustig2007sparse}, and more.
The problem can be formalized as
\begin{equation}
\label{eq:l0}
\min \|\bold x\|_0 \ \ \text{ s.t. } \ \ \bPhi \bold x = \bold y,
\end{equation}
where $\bPhi \in \R^{n \times m}$ is a matrix with potentially more columns than rows,
$\bold y$ is the observation we are trying to represent using an unknown sparse vector $\bold x$, and
$\| \bold x \|_0$ stands for the number of non-zero elements in $\bold x$.
A column $\bvarphi_i$ of $\bPhi$ 
is variously referred to as a feature or an atom.
This problem has been studied extensively, resulting in myriad 
existing methods and algorithms.

In the present work, 
we study two seemingly disparate techniques for solving problem \eqref{eq:l0}:
Sparse Bayesian Learning (SBL) and Stepwise Regression.
SBL is based on the Automatic Relevance Determination (ARD) framework \citep{mackay1992bayesian, mackay1992evidence}
and is concerned with a generative model of the form
\begin{equation}
\label{eq:sblreg}
\bPhi \bold x = \bold y +  \boldsymbol \epsilon,
\end{equation}
where $\bPhi$ is a deterministic matrix, 
$\epsilon_i \sim \N(0, \sigma)$ are independent noise variables,
and the prior distribution over the coefficients is
$x_i \sim \N(0, \gamma_i)$ \citep{tipping2001sparse}.
Each coefficient has an independent prior variance $\gamma_i$,
the defining characteristic of the ARD prior.
Sparse Bayesian models are usually trained via type-II maximum likelihood estimation, which is the maximization of the marginal likelihood
with respect to $\boldsymbol \gamma$.
The logarithm of the marginal likelihood is
\begin{equation}
\begin{aligned}
\label{eq:nlml}
 \mathcal{L}(\boldsymbol \gamma) 
 &\eqdef \log \int p(\bold y | \bold x) p(\bold x| \boldsymbol \gamma) d \bold x \\
 &= -\by^* \bCi \by - \log | \bC | - n \log (2\pi),
 \end{aligned}
\end{equation}
where $\bC \eqdef (\sigma^2 \bold I + \bPhi \boldsymbol \Gamma \bPhi^*)$,
and $\boldsymbol \Gamma \eqdef \text{diag}(\boldsymbol \gamma)$.
Since every coefficient has its own prior variance,
the optimization effectively prunes extraneous features of $\bPhi$
if their prior variances approache zero.
Since this provably happens frequently \citep{wipf2005l0},
ARD and SBL are powerful tools for
promoting sparsity
 in a variety of applications.

Stepwise Regression, on the other hand,
is a class of well-known greedy algorithms that
add and delete features from a candidate solution 
based on two rules:
\begin{equation}
\label{eq:stepwise}
\arg \min_{i \not \in \act} \| \bold r_{\act \cup i} \|_2, 
\qquad 
\text{and}
\qquad
\arg \min_{i \in \act} \| \bold r_{\act \backslash i} \|_2,
\end{equation}
where $\bold r_\act$ is the least-squares residual 
using only features indexed by $\act$.
The intuitive appeal and practical performance of these greedy heuristics
have led separate communities to rediscover the same algorithms,
leading to a bewildering number of names.
The algorithm that selects features based on the left side of \eqref{eq:stepwise}
is known as Forward Regression, Forward Selection \citep{miller2002subset}, 
Order-Recursive Matching Pursuit (ORMP) \citep{cotter1999orls},
Optimized Orthogonal Matching Pursuit \citep{rebollo2002optimized},
and Orthogonal Least-Squares \citep{cotter1999orls}.
The algorithm that eliminates features based on the right side of \eqref{eq:stepwise}
is known as Backward Regression, Backward Elimination \citep{miller2002subset}, 
and Backward Optimized Orthogonal Matching Pursuit \citep{andrle2004boomp}.
Unless otherwise noted, we will refer to them simply as the forward and backward 
algorithm, respectively.

\paragraph{Contributions} 
We propose Relevance Matching Pursuit (\RMP), an algorithm which simultaneously maximizes \eqref{eq:nlml} locally,
and exhibits a surprising relationship to Stepwise Regression.
Indeed, by analyzing \RMP \ in the noiseless limit in Section~\ref{sec:rmp},
we derive \RMPZ, a combination of the well-known 
forward and backward algorithms.
Having established this connection, we derive novel theoretical guarantees
for Stepwise Regression on noisy data in Section~\ref{sec:stepwise}.
The guarantees for Forward Regression improve on existing 
deterministic and probabilistic results for Orthogonal Matching Pursuit.
The bounds of the tolerable perturbation for Backward Regression are,
to our knowledge,
the first that can be computed in polynomial time and provide an important insight:
the backward algorithm returns the {\it optimal} solution to the subset selection problem 
on a determined linear system,
as long as the residual of the solution is bounded by a quantity that is proportional to the
smallest singular value of the matrix.
We experimentally verify our prediction that \RMPZ \ 
exhibits comparable support recovery performance to \RMP \
and compare against 
some of the most potent feature selection
algorithms\footnote{Code made available at \href{https://github.com/SebastianAment/CompressedSensing.jl}{\texttt{CompressedSensing.jl}}.}
in Section~\ref{sec:experiments}.
The results demonstrate RMP's 
combination of performance and efficiency,
corroborating our theoretical analysis. 
The only method with a consistent performance advantage over \RMP \ and \RMPZ \ is the ARD-based reweighted $l_1$-norm minimization of \citet{wipf2008},
though the approach is computationally substantially more expensive.


\section{Relevant Work}
Given the vast amount of work on sparsity-inducing methods, we focus 
on the most relevant to ours.
In the following, we will use the bra-ket notation $\bra \bold \cdot | \cdot \ket$ to denote an inner product.
A feature that has been selected by an algorithm is
considered active and 
we refer to the set $\act$ of all active features as the active set.

\paragraph{Basis Pursuit} 
Basis Pursuit (BP) is a framework for solving \eqref{eq:l0} via the following convex relaxation:
\begin{equation}
\label{eq:l1}
\min \|\bold x\|_1 \ \ \text{ s.t. } \ \ \bPhi \bold x = \bold y.
\end{equation}
Under certain assumptions on the matrix $\bPhi$
and the sparsity level $k = \| \bold x \|_0$,
\eqref{eq:l1} has the same global optimum as \eqref{eq:l0}
\citep{chen2001atomic, candes2006robust}.
An important modification to BP for noisy observations is Basis Pursuit Denoising (BPDN)
which replaces the equality constraint with $\| \bPhi \bold x - \bold y\| \leq \delta$,
where $\delta$ is an upper bound on the perturbation of the signal \citep{chen2001atomic, donoho2006bpwnoise}.
This is closely related to the least absolute shrinkage and selection operator (LASSO)
\citep{tibshirani1996regression, zhao2007stagewise, bach2008consistency}.
\citet{cawley2007sparse} studied this principle in the Bayesian framework,
 where the $l_1$-regularizer is equivalent to a Laplacian prior.
Other notable algorithms based on relaxations of \eqref{eq:l0} include FOCUSS, an iterative least-squares 
scheme \citep{focuss},
and a reweighted $l_1$-norm minimization algorithm proposed in \citep{candes2008enhancing},
both approximating an entropic regularization term.
While algorithms based on relaxations can offer strong sparse recovery guarantees and performance, they can be computationally expensive for large problems.
Efficient greedy algorithms have been developed for this reason.

\paragraph{Matching Pursuit}
An important family of greedy algorithms for \eqref{eq:l0}
are Matching Pursuit (MP) and its variants \citep{mallat1993pursuit}.
Matching Pursuit updates a candidate solution
one element at a time.
The specific element $i$ is chosen by the rule
$\arg \max_i | \bra \bvarphi_i | \bold r \ket|$,
where $\bvarphi_i$ is the $i$th column of $\bPhi$ and $\bold r = \bold y - \bPhi \bold x$ is the current residual. 
Orthogonal Matching Pursuit (OMP),
also known as Stagewise Regression,
 uses the same rule to add features,
but additionally optimizes all coefficients of the active set in each iteration:
\begin{equation}
\label{eq:omp}
\arg \max_i | \bra \bvarphi_i | \bold r_\act \ket|,
\end{equation}
where $\bold r_{\act}$ is the least-squares residual given the set of atoms $\act$.
Remarkably, OMP has theoretical guarantees for the problem of recovering the support of exactly sparse signals, even for noisy measurements~\citep{davis1997adaptive, greedisgood, tropp2007recovery, rangan2009orthogonal, cai2010noise}.
Recently, Matching Pursuits have served as inspiration for optimization algorithms:
\citet{tibshirani2015general} proposed a general framework for stagewise algorithms
with applications to group-structured learning, matrix completion, and image denoising.
\citet{pmlr-v80-locatello18a} developed a unified analysis of MP and coordinate ascent algorithms and \citet{combettes2019blended} proposed Blended Matching Pursuit,
combining coordinate descent and gradient steps 
to compute sparse minimizers of general convex objectives quickly.

\paragraph{Stepwise Regression}

We here focus on known theoretical guarantees of the forward and backward algorithms
and existing algorithms that combine them.
\citet{das2018approximate} 
proposed a notion of approximate submodularity 
and showed that it is satisfied by the coefficient of determination, $R^2$.
In this way, they proved approximation guarantees of the forward algorithm and OMP to the optimal solution of the subset selection problem,
which were generalized by \citet{elenberg2018restricted} for general convex objectives.
\citet{backwardoptimality} analyzed the backward algorithm
and proved the existence of a bound on the perturbation magnitude 
that guarantees the recovery of the support of sparse solutions of linear systems.

\citet{andrle2004boomp} proposed running the forward and backward algorithm consecutively, but did not provide theoretical guarantees, nor empirical comparisons against other algorithms.
\citet{nipsfoba2009} proposed FoBa, which combines both forward and backward heuristics 
 into an adaptive algorithm.
Similarly, \citet{rao2015forward} proposed a
 forward-backward algorithm for the optimization of convex relaxations of \eqref{eq:l0}
based on atomic norms
and, most recently, \citet{borboudakis2019forward} 
proposed an early-dropping heuristic
for forward-backward algorithms
for general feature selection problems.

\paragraph{Sparse Bayesian Learning}
The first algorithms for the optimization of \eqref{eq:nlml} for SBL
were based on expectation-maximization (EM) updates 
and the fixed-point updates of MacKay \citep{tipping2001sparse}.
Though these methods are able to obtain sparse solutions to \eqref{eq:sblreg},
they have no convergence guarantees and are slow for large problems,
due to the at least quadratic scaling with the number of features \citep{tipping2001sparse}.
\citet{wipf2004sparse} showed how to adapt the EM-based SBL algorithm
to the $l_0$-minimization problem \eqref{eq:l0} and proved that, in contrast to BP, the resulting optimization problem has the same global optimum as \eqref{eq:l0}
and suffers from fewer local minima than competing non-convex relaxations.
 Subsequently, \citet{wipf2008} 
 showed that the usual type-II approach can be interpreted as a type-I (MAP) approach with a special non-factorial prior.
 Using this insight, they proposed an algorithm based on reweighted $l_1$-norm
minimization, which provably converges to a local maximum of the marginal likelihood, 
performs at least as well as BP in recovering sparse signals,
 and usually outperforms techniques based on $l_1$, $l_2$, and entropy regularization  \citep{wipf2009sparse},
especially when dictionaries are structured and coherent \citep{wipf2011sparse}.
 
\section{Relevance Matching Pursuit}
\label{sec:rmp}

This section first recapitulates the derivation of the coordinate ascent updates
for SBL derived by \citet{tipping2003fast},
subsequently introduces the Relevance Matching Pursuit algorithm,
and analyzes the algorithm's behavior as the 
noise variance approaches zero.

\subsection{SBL via Coordinate Ascent}

Recall from the introduction that 
$\bC = (\sigma^2 \bold I + \bPhi \boldsymbol \Gamma \bPhi^*)$
is the covariance of the marginal distribution and
$\boldsymbol \Gamma = \text{diag}(\boldsymbol \gamma)$
is the prior variance of the weights $\bold x$.
In the context of SBL,
we refer to $\act = \{ i | \gamma_i \neq 0 \}$ as the active set.
Following the analysis of \citet{tipping2003fast}
and based on the Woodbury matrix identity,
we separate out the contribution of a single prior variance $\gamma_i$ to the marginal likelihood \eqref{eq:nlml}:
\[
\begin{aligned}
\mathcal{L}(\bgamma) 
&= \mathcal{L}(\bgamma_{-i}) + \frac{1}{2} \left( \frac{q_i^2} {\gamma_i^{-1} + s_i} - \log \frac{1}{1+ \gamma_i s_i}\right) \\
&=\mathcal{L}(\bgamma_{-i}) + l(\gamma_i),
\end{aligned}
\]
where 
$q_i \eqdef \bvarphi_i \bold C_{\act \backslash i}^{-1} \bold y$
and $s_i \eqdef \bvarphi_i \bold C_{\act \backslash i}^{-1} \bvarphi_i$, 
also termed the "quality" and "sparsity" factors by \citet{faul2002analysis}.
$\bold C_{\act \backslash i}$ is as in \eqref{eq:nlml} 
but only includes the features $\bvarphi_k$ and 
corresponding prior variances $\gamma_k$ for $k \in \act \backslash i$. 
Crucially, the argument of the maximum of the marginal likelihood with respect to a single prior variance $\gamma_i$ is unique and has a closed form:
\begin{equation}
\label{eq:bestgamma}
\arg \max_{\gamma_i} \ l(\gamma_i) = 
\begin{cases} (q_i^2 - s_i) / s_i^2 & q_i^2 > s_i \\
 0 & \text{else} 
\end{cases}.
\end{equation}
Equation \eqref{eq:bestgamma} is the basis of the efficient coordinate ascent updates 
put forth in \citet{tipping2003fast}.
Associated with each coordinate update is a change in the marginal likelihood.
If $q_i^2 > s_i$,
we denote by
 $\Delta_{\text{add}}(i)$,
 respectively $\Delta_{\text{update}}(i)$,
 the change in the marginal likelihood corresponding to 
 setting a $\gamma_i$ which was previously zero, respectively non-zero, via equation \eqref{eq:bestgamma}.
 
We now make two preliminary observations.
Given a subset of features $\act$, and a noise variance $\sigma^2$,
the posterior mean $\bmu_{\act, \sigma}$ and variance $\bSigma_{\act, \sigma}$ 
of the subset of weights $\bold x_\act$ are given by
\[
\begin{aligned}
\bSigma_{\act, \sigma} 
&= (  {\bGamma_\act}^{-1} + \sigma^{-2} \bPhi_\act^T \bPhi_\act)^{-1}, \\
\bmu_{\act, \sigma} 
&= \sigma^{-2} \bSigma_\act \bPhi_\act^* \bold y,
\end{aligned}
\]
where 
$\bPhi_\act$ and
$\bGamma_\act$
are the submatrices of $\bPhi$ and $\bGamma$ 
corresponding to $\act$.
The Woodbury identity gives
\[
\begin{aligned}
\bold R_{\act, \sigma}  
&\eqdef  \sigma^2 \bold C^{-1}
= \bold I - \bPhi_\act [\sigma^{-2} \bSigma_{\act}] \bPhi_\act^*, \\
\end{aligned}
\]
and thus,
$
\bold r_{\act, \sigma} 
\eqdef \bold y - \bPhi_\act \bmu_{\act, \sigma} 
= \bold R_{\act, \sigma} \bold y.$
We can now express the following result
on the condition which
leads SBL to include or exclude a feature.

\begin{lem}
\label{lem:selectiondeletion}
The optimum of the marginal likelihood with respect to $\gamma_i$ 
occurs at a non-zero value 
if and only if 
\[
| \bra \tilde \bvarphi_i | \bold r_{\act \backslash i, \sigma} \ket| > \sigma,
\]
where $\tilde \bvarphi_i \eqdef \bvarphi_i / \| \bvarphi_i \|_{\bold R_{\act \backslash i, \sigma}}$, 
$\| \bvarphi_i \|_{\bold R_{\act \backslash i, \sigma}}$
is the energetic norm  $(\bvarphi_i^*\bold R_{\act \backslash i, \sigma} \bvarphi_i)$ of $\bvarphi_i$, and $\act$ is the active set.
\end{lem}
Lemma \ref{lem:selectiondeletion} 
shows the direct and proportional dependency of the
acquisition and deletion conditions
on $\sigma$.
The following result characterizes 
the inactive feature ($\gamma_i = 0$) that
leads to the maximal increase in the marginal likelihood
upon its addition to the model.

\begin{lem}
\label{lem:rmp}
Let $\Delta_{add}(i)$ be the change in the marginal likelihood upon 
setting an inactive feature's prior variance $\gamma_i$ to its optimal value
via equation \eqref{eq:bestgamma}.
Then
\[
\arg \max_{i \not \in \act} \Delta_{add}(i)
= \arg \max_{i \not \in \act} | \bra \tilde \bvarphi_i | \bold r_{\act, \sigma} \ket|.
\]
where $\tilde \bvarphi_i = \bvarphi_i / \|\bvarphi_i\|_{\bold R_{\act, \sigma}}$, and $\act$ is the active set.
\end{lem}

By comparing the right side of the equation in Lemma \ref{lem:rmp}
to the acquisition criterion of OMP in \eqref{eq:omp},
we see that a feature selection strategy based on the maximal increase in the 
marginal likelihood is intimately related
to the family of Matching Pursuit algorithms.
This serves as the inspiration for the name of Relevance Matching Pursuit (RMP), described in the next section. 

\subsection{Algorithm Design}

In describing the coordinate ascent updates, \citet{tipping2003fast} purposefully left several choices open:
which variance $\gamma_i$ does the algorithm choose to update, add, or delete?
In which order should these operations proceed?
RMP arises from a particular choice for these design questions, enabling our analysis of the algorithm's behavior,
and proving to be closely related to Stepwise Regression.
The design principles give rise to Algorithm \ref{alg:rmp}
and are as follows:
\begin{enumerate}[1)]
\item Add features based on $\arg \max_{i \not \in \act} | \bra \tilde \bvarphi_i | \bold r \ket|$,
until there is no inactive feature with $| \bra \tilde \bvarphi_i | \bold r \ket|~>~\sigma$ left.
\item Remove features based on  $\arg \min_{i \not \in \act} | \bra \tilde \bvarphi_i | \bold r \ket|$, as long as there is a feature with $| \bra \tilde \bvarphi_i | \bold r \ket| \leq \sigma$.
\item Update the prior variance of the currently active atom
whose update leads to the largest increase in the marginal likelihood,
as long as there is a feature with $\Delta_{\text{update}}(i) > \delta_{\mathcal{L}}$,
where $\delta_\mathcal{L}$ is an input to the algorithm and
defines a convergence criterion.
\end{enumerate}

\begin{algorithm}[t]
\caption{Relevance Matching Pursuit \small (\RMP)}
\label{alg:rmp}
\begin{algorithmic}[1]
\STATE {\bfseries Input:} Dictionary $\bPhi$, signal $\bold y$, noise variance $\sigma^2$, convergence criterion $\delta_\mathcal{L}$
\STATE {\bfseries Ouput:} Support set $\act$, prior variances $\gamma_i$
\STATE Initialize $\act \gets \varnothing$
\STATE Initialize $\gamma_i \gets 0$ for all $i$\\
\WHILE {has not converged}
	\WHILE {$\exists i \not \in \act$ s.t. $| \bra \tilde \bvarphi_i | \bold r_{\act, \sigma} \ket | > \sigma$}
 		\STATE $i^* \gets \arg \max_{i\not \in \act} | \bra \tilde \bvarphi_i | \bold r_{\act, \sigma} \ket | $ \hfill \COMMENT{selection}
  		\STATE $ \act \gets \act \cup i^*$ \hfill \COMMENT{add to active set}
  		\STATE update $\gamma_{i^*}$ \hfill \COMMENT{set $\gamma_{i^*}$ via \eqref{eq:bestgamma}}
	\ENDWHILE
 	\WHILE {true}
   		\IF {$\exists i \in \act$ s.t. $| \bra \tilde \bvarphi_i | \bold r_{\act \backslash i, \sigma} \ket | \leq \sigma$ }
			\STATE $i^* \gets \arg \min_{i \in \act} | \bra \tilde \bvarphi_i | \bold r_{\act \backslash i, \sigma} \ket |$ 
  			\STATE $\act \gets \act \backslash i^*$ \hfill \COMMENT{elimination}
   		\ELSIF {$\exists i$ s.t. $\Delta_{\text{update}}(i) > \delta_{\mathcal{L}}$}
			\STATE $i^* \gets \arg \max_{i \in \act} \Delta \sub{update}(i)$ \hfill  \COMMENT{update}
   	 	\ELSE
 			\STATE break
		\ENDIF
 	\STATE update $\gamma_{i^*}$ \hfill \COMMENT{set $\gamma_{i^*}$ via \eqref{eq:bestgamma}} \\
	\ENDWHILE
\ENDWHILE
\end{algorithmic}
\end{algorithm}

In Algorithm \ref{alg:rmp}, we left the condition of the outer loop imprecise for a reason:
in addition to terminating after an improvement to the likelihood fails to exceed $\delta_\mathcal{L}$ and no feature is left to add or delete,
an implementation might include additional criteria, 
like a maximum runtime, number of iterations, or change in $\bgamma$.
We further note that the coordinate ascent updates to $\gamma_i$
do provably converge to a joint maximum,
not merely a stationary point \citep{faul2002analysis}.
These facts imply
\begin{lem}
 As $\delta_\mathcal{L} \to 0$, 
the $\bgamma$ returned by \RMP \ constitutes a local maximum of the marginal likelihood.
\end{lem}

\subsection{The Noiseless Limit} 
\label{sec:limit}

We now analyze the algorithm's behavior as the noise variance $\sigma^2$ approaches zero.
On a high level, this is analogous to the approach of \citet{wipf2004sparse, wipf2005l0}
who studied the noiseless limit of the EM-updates for SBL.
We make use of the following property of the Moore-Penrose inverse.
\begin{lem}
\label{lem:r0}
Assume the columns of $\bPhi_\act$, are linearly independent. Then
\[
\bold R_\act \eqdef 
(\bold I - \bPhi_\act \bPhi_\act^+)
= \lim_{\sigma \to 0^+} \bold R_{\act, \sigma}.
\]
\end{lem}
Note that $\bold r_{\act} = \bold R_\act \bold y $ is the ordinary least-squares residual of the active features $\bPhi_\act$ and $\bold y$.
The following technical result is crucial in establishing the connection 
of \RMP \ to Stepwise Regression.
\begin{lem}
\label{lem:ols}
Let $\bold r_\act$ be the least-squares residual associated with a feature set $\act$.
Then
\begin{equation}
\label{eq:ols}
\begin{aligned}
\| \bold r_\act \|_2^2 - \|\bold r_{\act \cup i} \|_2^2
&= | \bra \bvarphi_i, \bold r_\act \ket |^2 / \| \bvarphi_i\|_{\bold R_\act}^2, \text{and}\\
\| \bold r_{\act \backslash i} \|_2^2 - \|\bold r_{\act} \|_2^2
&= | \bra \bvarphi_i, \bold r_\act \ket |^2 / \| \bvarphi_i\|_{\bold R_{\act\backslash i}}^2.
\end{aligned}
\end{equation}
\end{lem}

An immediate corollary of Lemma \ref{lem:ols} is that
$\arg \min_i \|\bold r_{\act \cup i} \| = \arg \max_i  | \bvarphi_i \cdot \bold r_\act | / \| \bvarphi_i\|_{\bold R_\act}$,
and a similar expression for the second equation in \eqref{eq:ols}.
As Lemma \ref{lem:r0} implies that
the addition criteria of \RMP \
converge to the right-hand side of this expression,
the criteria in fact converge to the ones of 
the forward and backward algorithm in equation \eqref{eq:stepwise}.

It remains to study what happens to the prior-variance-update step 
in line 16 of Algorithm \ref{alg:rmp}.
Notably, $\bold R_\act$ is independent of $\bGamma$ under the assumption
of Lemma \ref{lem:r0}, and
thus, so are the addition and deletion criteria of \RMP \ in this limit.
Therefore, updating and keeping track of $\bGamma$ is irrelevant
for the execution of the algorithm if the active set is linearly independent.

This observation gives rise to Algorithm \ref{alg:rmp0}, which we term \RMPZ,
which models \RMP \ when there are no linearly dependent column in the active set.
This is not a restrictive assumption since
\RMPZ \ starts with an empty $\act$ and stops adding 
columns when a feasible solution is found since then $\bold r_\act = 0$,
which happens at the latest when $n$ columns are added.
Inspired by the acquisition criterion in Lemma~\ref{lem:selectiondeletion},
we introduce an acquisition and deletion threshold $\delta$ separate from $\sigma$
to Algorithm \ref{alg:rmp0}, which makes the algorithm capable of handling noise.
Setting $\delta \leftarrow 0$ corresponds to the noiseless limit of \RMP.

\begin{algorithm}[t]
 \caption{\RMPZ}
 \label{alg:rmp0}
 \begin{algorithmic}[1]
\STATE {\bfseries Input:} Dictionary $\bPhi$, signal $\bold y$, tolerance $\delta$
\STATE {\bfseries Result:} Support set $\act$
\STATE Initialize $\act \gets \varnothing$
\WHILE {has not converged}
	 \WHILE {$\exists i \in \act$ s.t. $\|\bold r_{\act} \|^2 - \| \bold r_{\act\cup i} \|^2 > \delta^2$}
		\STATE $i^* \gets \arg \max_{i\not \in \act} \| \bold r_{\act \cup i} \|$ \COMMENT{selection}
  		\STATE $ \act \gets \act \cup i^*$ 
	\ENDWHILE
	
	\WHILE {$\exists i \in \act$ s.t. $\|\bold r_{\act \backslash i} \|^2 - \| \bold r_\act \|^2 \leq \delta^2$}
		\STATE $i^* \gets \arg \min_{i \in \act} \| \bold r_{\act \backslash i} \|$ \COMMENT{elimination}
  		\STATE $\act \gets \act \backslash i^*$
 	\ENDWHILE
\ENDWHILE
\end{algorithmic}
\end{algorithm}

\section{Stepwise Regression}
\label{sec:stepwise}
Having exposed a connection of 
Relevance Matching Pursuit to 
Stepwise Regression, we now provide 
novel theoretical insights about the forward and backward algorithms.
To this end, 
we briefly introduce necessary theoretical tools.

\subsection{Theoretical Preliminaries}
Prior work has established that the performance of many feature selection
and sparse recovery algorithms 
is highly dependent on the correlation of different features.
The following definition quantifies this notion.
\begin{defi}[Coherence]
The coherence $\mu$ of a matrix $\bPhi$, whose columns have unit norm, is defined as
\[
\mu \eqdef \max_{i \neq j} | \bra \bvarphi_i | \bvarphi_j \ket |.
\]
\end{defi}
The coherence is a measure of the orthogonality of $\bPhi$.
It can lead to pessimistic estimates, as it only considers the maximal inner product of two columns.
\citet{greedisgood} introduced the Babel function
to generalize the coherence
by measuring the maximal sum of absolute inner products between a
column and {\it a set} of columns.
Its name is inspired by the Tower of Babel, 
since the function measures "how much
the atoms are speaking the same language",
and it can be used to derive sharper results.

\begin{defi}[Babel Function]
The Babel function $\mu_1$ of a dictionary $\bPhi$ is defined as
\[
\mu_1(k) \eqdef \max_{|\mathcal{I}| = k} \max_{i \not \in \mathcal{I}} \ \sum_{j \in \mathcal{I}} \ | \bra \bvarphi_i | \bvarphi_j \ket |.
\]
Notably, $\mu_1(1) = \mu$ and $\mu_1(k) \leq \mu k$ \citep{greedisgood}.
\end{defi}

Using this notion,
\citet{greedisgood} proved that a necessary and sufficient condition 
for Orthogonal Matching Pursuit to recover {\it any} $k$-sparse vector
is the Exact Recovery Condition.

\begin{thm}[\citet{greedisgood}]
\label{thm:erc}
Orthogonal Matching Pursuit and Basis Pursuit succeed in recovering the support $\supp$ of a $k$-sparse $\bold x$ from $\bold y = \bPhi \bold x$ if
\[
\max_{j \not \in \supp} \ \| \bPhi_\supp^+ \bvarphi_j \|_1 < 1.
\]
Further, this holds if $\mu_1(k) < 1/2$.
\end{thm}

\citet{soussen2013joint} used the connection of the forward algorithm and OMP
that was exposed through Lemma \ref{lem:ols} to jointly analyze the two algorithms,
proving that the exact recovery criterion 
in Theorem \ref{thm:erc} is necessary and sufficient
for both algorithms to retrieve {\it any} sparse signal with no noise.
In particular,
for each algorithm, there is a sparse signal  which cannot be recovered in $k$ steps,
if the inequality doesn't hold.
In this sense, the recovery guarantee for the forward algorithm without noise cannot be improved.
\citet{greedisgood} further points out
that even the condition on the Babel function is necessary
for exact recovery.

\subsection{Forward Regression}

For the results in this subsection, assume that $\bPhi$ has $l_2$-normalized columns.
In comparison to the work of \citet{cai2010noise} on OMP with noise,
our main theoretical contributions for the analysis of the forward algorithms
are related to the necessary recovery condition
on the Babel function, $\mu_1(k) < 1/2$,
and its implications.
Our analysis leads to both tighter deterministic
and probabilistic bounds on the tolerable noise magnitude.
\begin{thm}
\label{thm:ercn}
Orthogonal Matching Pursuit and Forward Regression recover the support of a $k$-sparse vector
in $k$ iterations
 provided
the Babel function $\mu_1$ 
and the perturbation $\beps$ of the target $\bold y$ satisfy
\[
\frac{1 - 2\mu_1(k)}{\sqrt{2[1+\mu_1(k)]}}
 \min_{i \in \supp} |x_i| \geq  \| \beps \|_2.
\]
\end{thm}

Theorem \ref{thm:ercn} allows for a non-zero amount of noise
as long as $\mu_1(k) < 1/2$, which is necessary even in the noiseless case.
The tolerable magnitude increases with decreasing Babel function value,
and is proportional to the magnitude of the smallest non-zero entry in $\bold x$.
Compared to Thm. 10 of \citet{bruckstein2009sparse} and Thm.~1 of \citet{cai2010noise},
Thm. \ref{thm:ercn} applies to both OMP and Forward Regression~(FR)
and improves the factor of $2$ in both prior results to $\sqrt{2(1+\mu_1(k))}$.
This allows up to around $40\%$ more noise while generalizing the results, and converges to $\sqrt{2}$ as $\mu \to 0$,
the provably tightest constant even for orthogonal matrices.
We also derived the following novel probabilistic guarantee for perturbations that are
normally distributed.

\begin{thm}
\label{thm:percn}
Suppose
$\beps \sim \N(0, \sigma^2\bold I_n)$,
and let $\delta = [1/2-\mu_1(k)] \min_{i \in \supp} |x_i| / \sigma > 0 $ and $d = \lceil m/k \rceil$.
Orthogonal Matching Pursuit and Forward Regression recover the support of 
a $k$-sparse signal with probability exceeding
\[
1 - \left \lceil \frac{m}{k} \right \rceil 
\left(\frac{1+\kappa_1(k)}{1-\mu_1(2k)}\right)^{k/2} \left(1- \erf(\delta / \sqrt{2\kappa_1(k)}) \right)^k.
\]
where $\kappa_1(k) \eqdef (1+\mu_1(2k)) /(1-\mu_1(k))$.
For $\mu_1(2k)<1/2$, the result further holds with probability exceeding 
\[
1 - 
\left \lceil \frac{m}{k} \right \rceil 
\left( \frac{4}{\sqrt{\pi}\delta} e^{- \delta^2/ 6} \right)^k.
\]
\end{thm}

\begin{figure}[h!]
\centering
\includegraphics[width=\columnwidth]{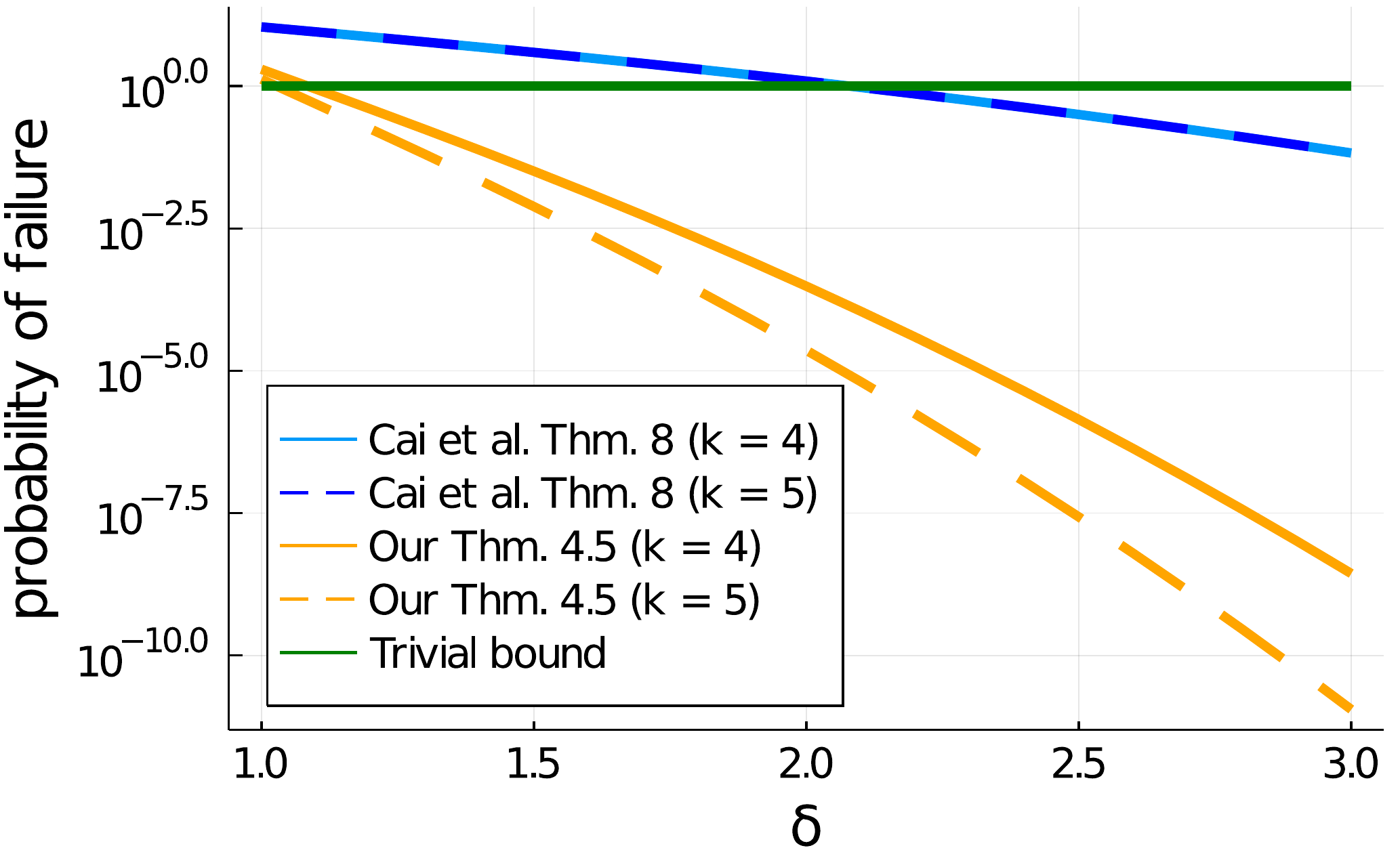}
\caption{Comparing existing probabilistic bounds of \cite{cai2010noise} to Theorem~\ref{thm:percn}
for a matrix with $m = 16$ columns,
where $\delta = [1/2-\mu_1(k)] \min_{i \in \supp} |x_i| / \sigma$.}
\label{fig:probabilistic_bounds}
\end{figure}

By comparison, the existing bounds of \citet{ben2010coherence} 
and \citet{cai2010noise} 
bound the probability of failure by 
$2m e^{-\delta^2/2} / \delta$. 
Theorem~\ref{thm:percn} thus guarantees an {\it earlier} and much {\it sharper} phase transition (see Fig.~\ref{fig:probabilistic_bounds}). 
Critically, Theorem~\ref{thm:percn} makes use of the approximately isometric structure of subsets of columns of $\bPhi$, which is already inherent to the deterministic analysis, 
to derive a strong bound for $\| \bPhi^* \beps\|$,
for which prior works applied bounds for generic $\bPhi$.
This allows us to apply much stronger {\it multiplicative} bounds 
since $\bPhi_\act^* \beps$, an important quantity in the analysis,
has an approximately diagonal covariance.

\paragraph{Sparse Approximation and Exact Recovery} 
\citet{elenberg2018restricted} and \citet{das2018approximate}
provide elegant theoretical insights on the performance of OMP and FR.
Both works present {\it approximation} guarantees,
but do not provide {\it exact recovery} guarantees, as Theorems \ref{thm:ercn} and \ref{thm:percn}. 
The results are thus highly complementary; neither subsumes the other.
To elaborate, their results for the $R^2$ score guarantee that FR's result
is at most a factor of $(1-e^{-\gamma})$ from the optimal value,
where $\gamma = \sigma_{\min}^2(\bPhi_\supp)$.
This is a strong result because it holds generally, even for non-sparse vectors and arbitrary noise.
On the other hand, if $\gamma > 1/2$ and a sparse vector generated the target with small noise, then Theorem~\ref{thm:ercn} guarantees the exact recovery of the support,
while the approximation guarantee with $\gamma~\approx~1/2$ only ensures FR to explain $\approx 40\%$ of the target variance.

\subsection{Backward Regression}
We now state our main optimality result for the backward algorithm.
The existence of such a result was already proved in \citet{backwardoptimality},
though their bound is NP-hard to evaluate.
In contrast, Theorem \ref{thm:berc} reveals a proportional dependence 
of the tolerable noise on the smallest singular value of the matrix.
While intuitive, this is to our knowledge the first result that makes this intuition precise.

\begin{thm}
\label{thm:berc}
Suppose $\bPhi$ has full column rank.
Then Backward Regression recovers the support $\supp$ 
of a $k$-sparse $m$-dimensional $\bold x$ in $m-k$ iterations if 
\[
\frac{\sigma_{\min}(\bPhi)}{\sqrt{2[2-\sigma_{\min}(\bPhi)^2]}}
\min_{i \in \supp} | x_i | 
> \| \beps \|_2,
\]
where $\sigma_{\min}(\bPhi)$ is the smallest singular value of $\bPhi$.
\end{thm}
Remarkably, 
the Backward Regression~(BR) only requires linear independence of the columns,
which is implied by $\mu_1(k) < 1$,
but only applies to {\it determined} systems.
Despite this necessarily stronger assumption, Thm.~\ref{thm:berc} is the first efficiently evaluable guarantee {\it for this case}. 
While the underdetermined case is most challenging, SBL is popularly applied to a kernel regression model for which the corresponding system is determined (see Sec.~\ref{sec:regression}).

Still, because of this stronger assumption, it is important to connect the result 
for the backward algorithm with the already analyzed forward algorithm,
the combination of which does apply to underdetermined systems.
To this end, suppose a forward algorithm terminates with an arbitrary superset $\act$ of the true support, which is a weaker assumption than that of exact recovery.
Then Theorem \ref{thm:berc} applied to the submatrix $\bPhi_\act$
guarantees that the backward algorithm only deletes irrelevant features if the noise is not too large.
The following corollary formalizes this idea.
\begin{cor}
\label{cor:berc}
Suppose $\bPhi_\act$ has full column rank, $|\act| = k$,
and $\supp \subset \act$.
Then Backward Regression recovers the correct support $\supp$
in $|\act| - |\supp|$ iterations, provided
\[
\sqrt{\frac{1-\mu_1(k)}{2[1+\mu_1(k)]} }
\min_{i \in \supp} | x_i | 
> \| \beps \|_2.
\]
\end{cor}
Note the striking similarity of the bounds in Corollary \ref{cor:berc} and Theorem \ref{thm:ercn}, though the former is stronger.
As \citet{backwardoptimality} already established,
the backward algorithm is not only capable of recovering the support of an {\it exactly} sparse vector,
but in fact can solve the subset selection to optimality, 
provided the residual of the optimal solution is small enough.
That is, for an arbitrary $\bold y$, not necessarily generated by $\bPhi \bold x$
with a sparse $\bold x$, we have 
\begin{thm}
Let $\bold x_k$ be the vector that achieves the smallest residual norm $\|\bold y - \bPhi \bold x\|$ among all vectors $\bold x$ with $k$ or fewer non-zero elements.
If the associated residual $\bold r_k \eqdef \bold y - \bPhi \bold x_k$ satisfies the bound in Theorem~\ref{thm:berc} in place of $\beps$,
Backward Regression recovers $\bold x_k$,
or equivalently, solves the subset selection problem to optimality.
\end{thm}

In an earlier short paper, we provided a precursory result 
for the backward algorithm \citep{ament2021optimality}. \linebreak
The full results we present herein provide tighter bounds and connect to the forward algorithm via Corollary~\ref{cor:berc}. 
Similar to the results for the forward algorithm in the present work, the bound in Theorem~\ref{thm:berc} also converges to $\sqrt{2}$ as $\mu \to 0$, the provably tightest constant even for orthogonal matrices.

\begin{figure*}[t!]
\includegraphics[width=\textwidth]{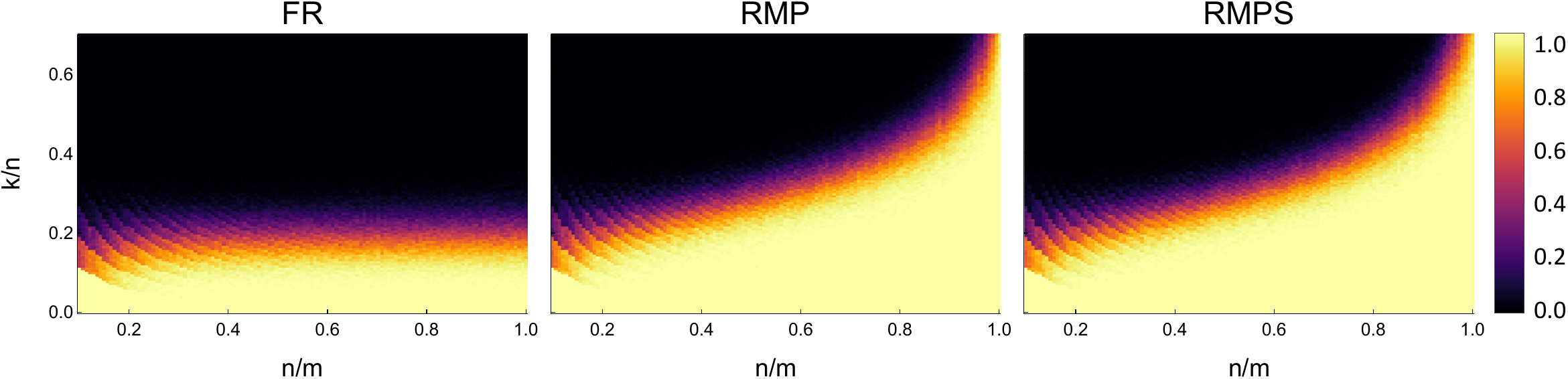}
\caption{Empirical frequency of support recovery as a function of the sampling
ratio $n/m$ and sparsity ratio $k/n$ for matrices with $m=128$ and $\| \beps \|_2 = 10^{-2}$ for Forward Regression (left), \RMPZ \ (middle) and \RMP \ (right).}
\label{fig:phase_transitions}
\end{figure*}

\section{Numerical Experiments}
\label{sec:experiments}
The preceding results are powerful in their own right. 
Further,
they guarantee that the backward stage of \RMPZ \ 
succeeds if
the forward stage terminates with a superset of the true support,
and the perturbation $\beps$ is not too large.
Therefore, we would expect RMP to have increasing 
support recovery performance as the sampling ratio $n/m$ increases.
The validation of this hypothesis is the goal of our first experiment.
Subsequently, we benchmark the recovery performance of a number of algorithms
on uncorrelated and correlated features with noise, a kernel regression task,
and end with a discussion of the experiments.

\subsection{Setup and Implementation} 

Beside RMP, we implemented
OMP, FR,
FoBa \citep{nipsfoba2009}, 
and the {\it steepest} coordinate ascent algorithm for SBL of \citet{tipping2003fast} (FSBL).
We compare two versions of \RMPZ \ in the experiments:
\RMPZ, and \RMPZ+.
The former denotes Algorithm \ref{alg:rmp0} 
with only one iteration of the outer loop,
while the latter terminates once the support stabilizes. 
We also compare against 
BPDN via constrained $l_1$-norm minimization,
and the SBL-based reweighted $l_1$-norm algorithm of \citet{wipf2008} (BP ARD).
We implemented all algorithms in {J}ulia~\citep{bezanson2017julia},
using the JuMP framework \citep{DunningHuchetteLubin2017}
to model the BP-approaches as second-order cone programs,
and solve them using ECOS \citep{bib:Domahidi2013ecos}
with default settings.
All experiments were run on a workstation 
with an Intel Xeon CPU X5670 and 47 GB of memory.

For the synthetic experiments, the weights $\bold x$ are random $k$-sparse vectors with $\pm 1$ entries and the targets $\bold y$ were perturbed by random vectors distributed
uniformly on the $10^{-2}$-hypersphere. 
For all algorithms, we input $\delta = 2\|\beps\|$ to simulate a small misspecification of the tolerance parameter that is likely to occur in practice. 
See also the supplementary materials for how \RMP \ could be adapted to infer the tolerance parameter, which however is a non-convex problem.
Note that the stopping criteria 
of some of the algorithms depend differently on $\delta$:
For OMP and BPDN, it is a constraint on the residual norm,
for \RMPZ, and FoBa, it is a bound on the marginal improvement 
in residual norm,
and for \RMP, we assign $\sigma \rightarrow \delta$
and note that the stopping criterion depends on the 
more complex expression of Lemma~\ref{lem:selectiondeletion}.

Since the BP cone-programs do not directly return sparse solutions,
we determine their support by dropping all entries below $\|\beps\|/ 10$.
In setting the threshold below the noise,
we highlight that standard BPDN introduces a bias,
while the ARD-based approach maintains the same global optimum as the $l_0$-minimization problem \citep{wipf2008}.
We stress that BP leads to sparse solutions in theory \cite{tropp2006relax},
but does not yield exactly sparse solutions using numerical LP-solvers,
which terminate with many coefficients very close, but not equal to zero,
necessitating the thresholding.

\subsection{Phase-Transitions}

First, we study the support recovery performance 
of Forward Regression (FR), \RMPZ \ and \RMP \
as a function of the sampling ratio $n/m$ of the matrix $\bPhi$
and the sparsity ratio $k/n$ of the weights $\bold x$.
Figure~\ref{fig:phase_transitions}
shows the empirical frequency of support recovery 
on column-normalized Gaussian random matrices 
with $m = 128$ columns.
Every cell is an average over 256 independent realizations of the experiment.

In accordance with the results for Stepwise Regression in Section \ref{sec:stepwise},
the recovery performance the RMP algorithms 
increases with the sampling ratio,
since the likelihood that the forward stage recovers a superset of the true support increases.
In contrast, the success of the forward algorithm in isolation  
is chiefly dependent on the sparsity ratio
and apparently independent of $n/m$.
The ridges in the bottom left of each plot are due to rounding effects 
since we set $k = \lceil (k/n) (n/m) m \rceil$.
Importantly, the performance of \RMPZ \ is virtually identical
to \RMP \,
and constitutes a first experimental validation of our analysis in Section~\ref{sec:rmp}.

\subsection{Support Recovery with Noise}

The following experiments have been established in the literature as a widespread proxy for performance on a variety of tasks, and thus allow for comparison to other reported results, for example those of \citet{wipf2004sparse}, \citet{candes2005error, candes2008enhancing}, and \citet{he2017bayesian}.
In particular, 
we record the empirical frequency of support recovery 
of a large set of algorithms as a function of the sparsity level $k$
for two types of matrices.
First, Gaussian random matrices and
second, matrices generated as
$\bPhi = \sum_p \frac{1}{p^2} \bold u \bold v^*$,
where $\bold u, \bold v$ have standard normal entries,
inspired by the experiments in \citet{xin2016maximal}.
The two types of matrices exhibit low and high column correlations, respectively.
We used matrices of size 64 by 128
and $l_2$-normalized the columns.
The results reported in Tables \ref{tbl:uncorrelated}
and \ref{tbl:correlated} are averages over 1024 independent realizations
with a 95\% confidence interval below 0.02.

\begin{table}[t]
  \caption{Recovery probability for {\it uncorrelated} features}
  \label{tbl:uncorrelated}
  \centering
\begin{small}
\begin{tabular}{llllllllll}
 \toprule
      & & \multicolumn{4}{c}{Sparsity}           \\
    \cmidrule(r){3-6}
Type & Algorithm & 12 & 16 & 20 & 24\\ 
\midrule
\multirow{5}{*}{MP} & OMP & $0.53$ & $0.15$ & $0.02$ & $0.00$ \\
 & FR & $0.54$ & $0.14$ & $0.01$ & $0.00$ \\
& FoBa & $\bf 0.99$ & $\bf 0.82$ & $\bf 0.31$ & $\bf 0.04$ \\
& {\bf RMP$_0$} & $\bf 0.99$ & $0.80$ & $\bf 0.31$ & $\bf 0.04$\\
& {\bf RMP$_0$+} & $\bf 0.99$ & $0.80$ & $\bf 0.31$ & $\bf 0.04$\\
\midrule[.25pt]
\multirow{2}{*}{SBL} & FSBL & $\bf 0.99$ & $0.80$ & $\bf 0.31$ & $\bf 0.04$ \\
& {\bf \RMP} & $\bf 0.99$ & $\bf 0.81$ & $\bf 0.31$ & $\bf 0.04$ \\
\midrule[.25pt]
\multirow{2}{*}{BP} & BP & $0.26$ & $0.05$ & $0.01$ & $0.00$\\
& BP ARD & $\bf 1.00$ & $\bf 1.00$ & $\bf 0.97$ & $\bf 0.70$ \\
\bottomrule
\end{tabular}
\end{small}
\end{table}

\begin{table}[t]
  \caption{Recovery probability for {\it correlated} features}
  \label{tbl:correlated}
  \centering
\begin{small}
\begin{tabular}{llllllllll}
 \toprule
      & & \multicolumn{4}{c}{Sparsity}           \\
    \cmidrule(r){3-6}
Type & Algorithm & 2 & 3 & 4 & 5\\ 
\midrule
\multirow{5}{*}{MP} & OMP & $0.00$ & $0.00$ & $0.00$ & $0.00$ \\
 & FR & $0.04$ & $0.01$ & $0.00$ & $0.00$ \\
& FoBa & $0.71$ & $0.46$ & $0.28$ & $\bf 0.17$\\
& {\bf RMP$_0$} & $\bf 0.72$ & $0.45$ & $0.28$ & $0.14$\\
& {\bf RMP$_0$+} & $\bf 0.72$ & $\bf 0.48$ & $\bf 0.32$ & $\bf 0.17$\\
\midrule[.25pt]
\multirow{2}{*}{SBL} & FSBL & $0.76$ & $0.54$ & $0.38$ & $0.26$ \\
& {\bf \RMP} & $\bf 0.81$ & $\bf 0.58$ & $\bf 0.45$ & $\bf 0.30$ \\
\midrule[.25pt]
\multirow{2}{*}{BP} & BP & $0.02$ & $0.00$ & $0.00$ & $0.00$\\
& BP ARD & $\bf 0.96$ & $\bf 0.91$ & $\bf 0.83$ & $\bf 0.70$ \\
\bottomrule
\end{tabular}
\end{small}
\end{table}

\begin{figure}
\centering
\includegraphics[width = .9\columnwidth]{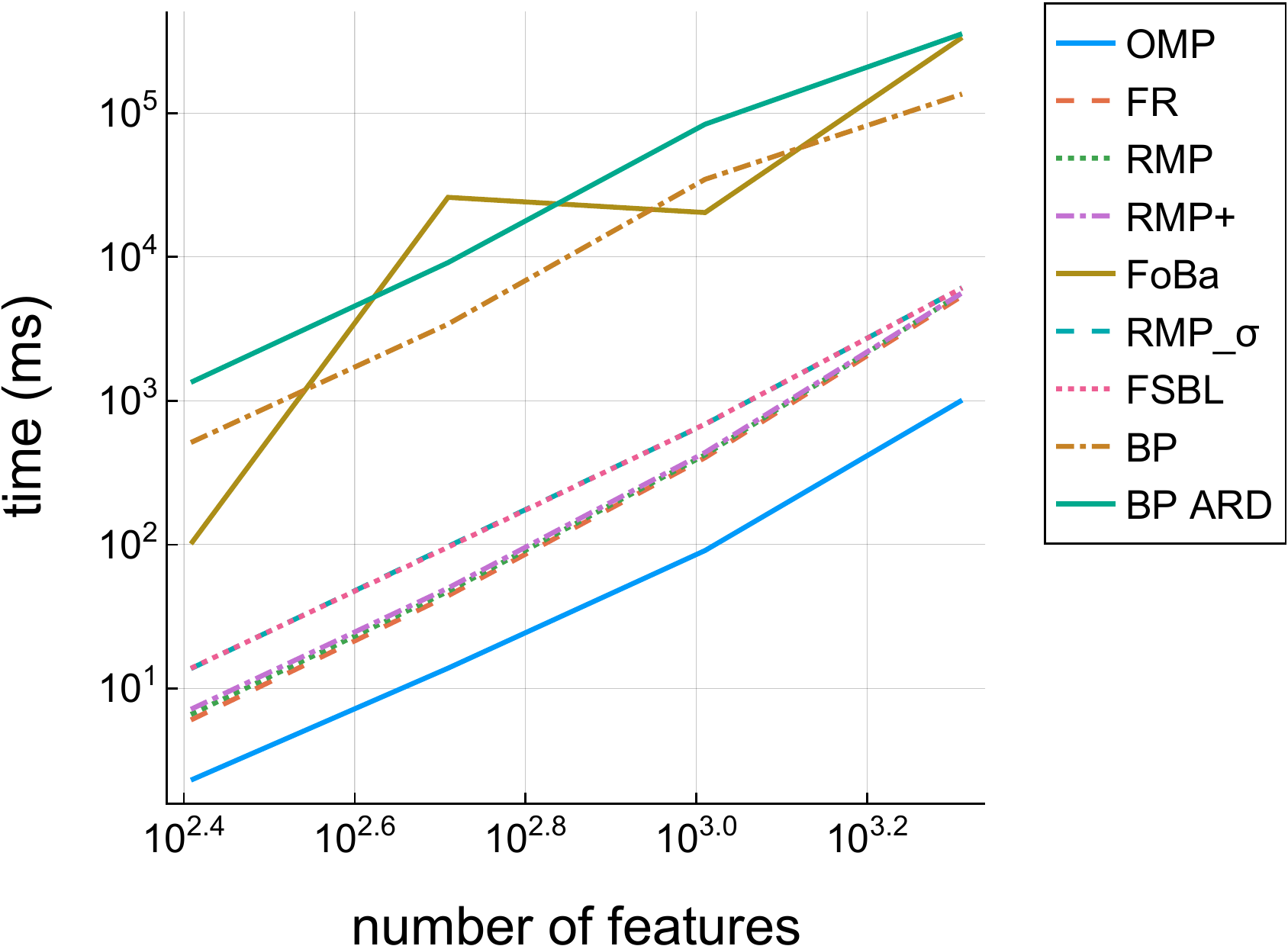}
\caption{Timings for several algorithms on matrices with a fixed
sampling ($1/2$) and sparsity ratio ($1/4$).}
\label{fig:time}
\end{figure}

BP ARD demonstrates the best recovery performance among all tested methods,
followed by \RMP.
Importantly,
\RMP \ exhibits similar performance to \RMPZ \
and virtually identical for uncorrelated features.
The differences between 
\RMPZ \ and FoBa are marginal and plausibly attributable 
to statistical fluctuations.
However, the differences in recovery performance of 
\RMP \ and FSBL is statistically significant for correlated features.
BP performs poorly since our experiment was designed to expose that it
does not preserve the same optimum as $l_0$-minimization.

Another notable approach 
is that of \citet{koyejo2014prior}, who proposed a greedy information-projection algorithm with applications to sparse estimation problems.
In our study, the algorithm performed well on Gaussian matrices but deteriorated 
similarly to OMP and FR on coherent matrices, which is expected since it solely takes forward steps.

Figure \ref{fig:time} shows runtimes
of the algorithms for matrices of increasing sizes.
As $m$ increases, we kept the ratios $n/m = 1/2$ and $k/m = 1/4$.
BP ARD is most time-consuming.
RMP and RMP$_0$ are on average two orders of magnitude faster.
The main performance difference between \RMP \ and \RMPZ \ 
comes from \RMP's $\gamma_i$-update,
which it can execute many times before converging.
FoBa should in principle scale comparably to \RMPZ, 
but doesn't, as we chose not to let it 
take advantage 
of the efficient backward updates
discussed in \citet{reeves1999efficient},
but not mentioned in \citet{nipsfoba2009},
which highlights their importance.
Last, a limitation of the timings for BP are that we used generic LP solvers,
while specialized algorithms exist \citep{beck2009fista, perez2019bucket} that can accelerate BP approaches.
On the other hand, keeping the sparsity ratio $k/n$ fixed while growing $m$
is to the detriment of the greedy algorithms, which would need much fewer iterations if the sparsity level $k$ was fixed instead. 
In all, the timings are not designed to be fair but to illustrate the inevitable 
trade-off between performance and efficiency.

\subsection{Sparse Kernel Regression}
\label{sec:regression}

\begin{figure}[t]
\centering
\includegraphics[width=.87\columnwidth]{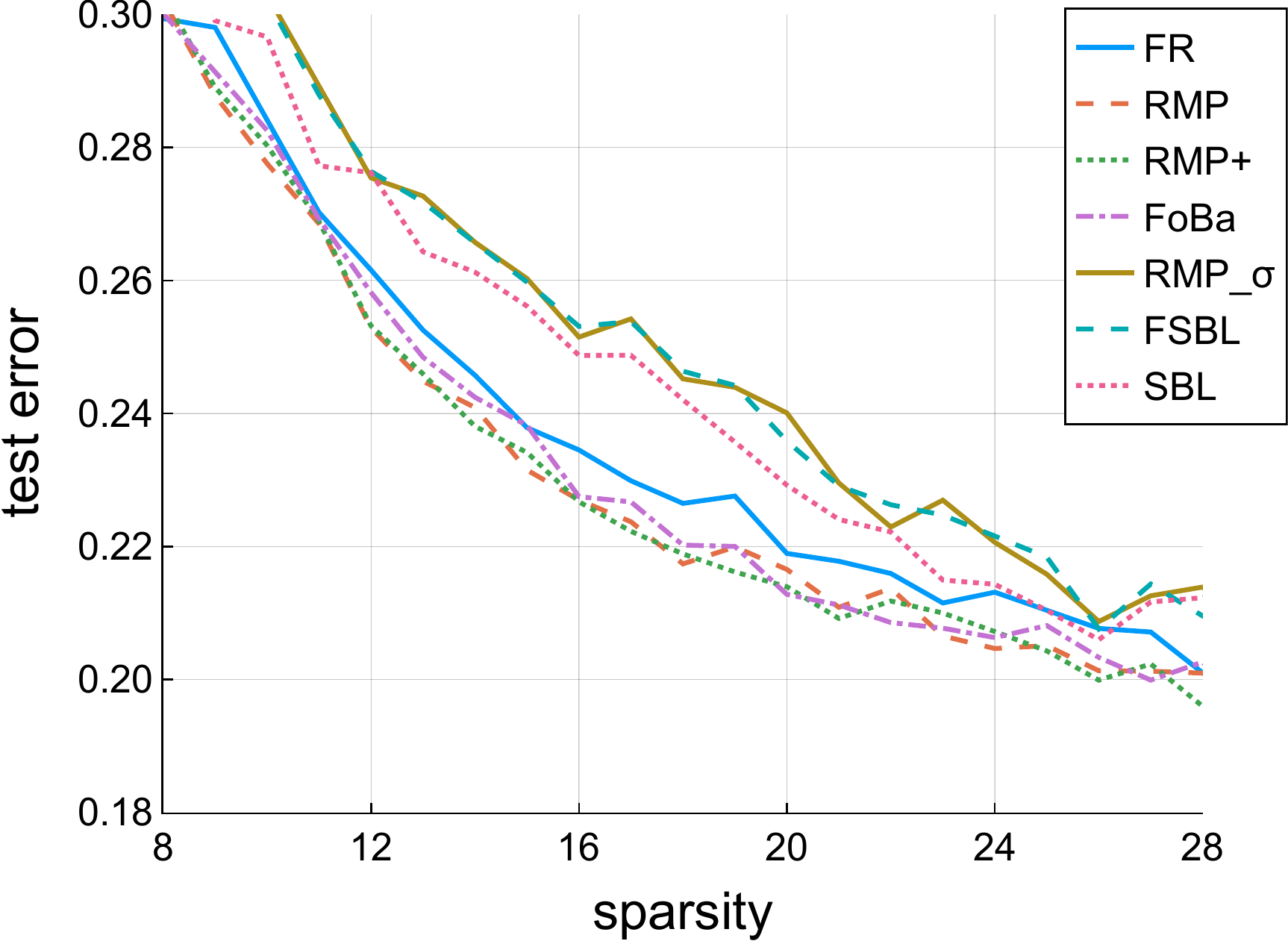}
\caption{
Test RMSE for a kernel regression model on the UCI Boston housing data
as a function of achieved sparsity.
}
\label{fig:boston}
\end{figure}

Following \citet{tipping2001sparse}, 
we apply the SBL-related algorithms to a kernel regression model.
In particular, given inputs $\{ \bold x_i\}$, we assume the responses are generated according to $y \sim \N(f(\bold x), \sigma^2)$, where
\begin{equation}
\label{eq:kernel}
f(\bold x) \ \eqdef \ \sum_i k(\bold x, \bold x_i) w_i,
\end{equation}
where $k$ is the Mat\'ern-3/2 kernel and weights~$w_i$.
Given a training set, we optimize $w_i$ using the SBL-related algorithms,
and evaluate on a test set using equation \eqref{eq:kernel}.
Figure~\ref{fig:boston} shows the mean test error as a function of sparsity on the UCI Boston housing data \citep{uci2017}, which contains 506 data points.
The results are averaged over 4608 sparsity-error values for each algorithm,
generated by evaluations for different tolerance parameters $\delta$ (i.e. $\sigma$) and random 75-25 train-test splits.
We conjecture that (F)SBL exhibits larger errors as it does not directly minimize squared errors, but instead the marginal likelihood.
While FR can be competitive for highly sparse solutions, it is not as effective as
RMP, RMP+ and FoBa, which achieve the best sparsity-error trade-off throughout all sparsity levels.

\section{Conclusion}

We proposed Relevance Matching Pursuit,
a coordinate ascent algorithm for SBL 
whose analysis reveals a surprising connection to Stepwise Regression.
The limiting algorithm \RMPZ \ closely tracks the performance of \RMP \ 
in our empirical evaluation, and is yet remarkably simple.
We provided
novel theoretical insights for Stepwise Regression, 
among them, an efficiently computable guarantee for the backward algorithm.
Our results further provide theoretical justification to practitioners using Stepwise Regression
in one of many statistics packages,
prominently in the widely-used SAS.
Finally, we hope these insights 
contribute to providing clarity in this vast space of the literature,
and inspire further research on powerful sparsity-inducing algorithms.

\small
\bibliography{SBLAndStepwise}

\end{document}